\documentclass[10pt]{article}

\usepackage[utf8]{inputenc}
\usepackage[T1]{fontenc}
\usepackage[margin=0.85in,top=0.7in,bottom=0.75in]{geometry}
\usepackage[protrusion=true,expansion=false]{microtype}
\usepackage{amsmath,amssymb}
\usepackage{booktabs}
\usepackage{array}
\usepackage{listings}
\usepackage{xcolor}
\usepackage{enumitem}
\usepackage{graphicx}
\usepackage[hidelinks]{hyperref}
\usepackage{titlesec}
\titlespacing*{\section}{0pt}{8pt}{4pt}
\titlespacing*{\subsection}{0pt}{6pt}{3pt}
\titlespacing*{\paragraph}{0pt}{5pt}{4pt}
\definecolor{codebg}{gray}{0.96}
\title{\textbf{A Numerically-Robust ROS\,2 Port of \mbox{iG-LIO}:\\
Diagnosing and Fixing Toolchain-Induced Failures in\\
Incremental GICP LiDAR-Inertial Odometry}}

\author{
Afonso E. Carvalho%
\thanks{Robotics Institute (RI), Carnegie Mellon University (CMU), Pittsburgh, USA.}
\thanks{Institute of Systems and Robotics (ISR), University of Coimbra (UC), Coimbra, Portugal.}%
\and
David Portugal\footnotemark[2]
\and
Paulo Peixoto\footnotemark[2]
}
\date{2026-07 \\[2pt] \small\textit{Technical Report}}

\begin{document}
\maketitle

\begin{abstract}
\noindent
\mbox{iG-LIO} is a tightly-coupled LiDAR-inertial odometry system fusing generalized-ICP and point-to-plane constraints in an iterated error-state Kalman filter over an incremental voxel map.
We report an open-source ROS\,2~Jazzy port of the original ROS\,1 implementation and, more importantly, the diagnosis of environment-induced numerical failures that appear only after the port: a mechanically faithful migration -- estimation mathematics left unchanged -- compiled and ran, yet diverged with \texttt{NaN} internal values.
Both causes trace to the modern ROS\,2 toolchain, not the algorithm: a Quality-of-Service (QoS) mismatch that silently drops and reorders IMU samples, and an uninitialized parallel-reduce accumulator arising from the \texttt{oneTBB}\,+\,Eigen combination shipped with current distributions.
We further correct Ouster point-field parsing to ensure correct point cloud undistortion with newer Ouster revisions, add Velodyne Velarray~M1600 support, provide both a compile-time-gated Livox \texttt{CustomMsg} path and a driver-free path for Livox sensors publishing standard \texttt{PointCloud2} (e.g.\ Mid-360), and expose the runtime via YAML.
The result has been validated in an Ouster OS0 Rev7, an Ouster OS1 Rev 7, and a Livox MID-360. This report is a citable reference for the port itself, not a claim on the underlying algorithm~\cite{chen2024iglio}.
The ROS\,2 port of iG-LIO described in this document can be found at \url{https://github.com/Forestry-Robotics-UC/ig_lio/tree/ros2-jazzy}.
\end{abstract}

% ===========================================================================
\section{Introduction}
% ===========================================================================
\mbox{iG-LIO}~\cite{chen2024iglio} is an incremental GICP-based, tightly-coupled LiDAR-inertial estimator, but its public implementation targets ROS\,1, now end-of-life.
Porting to ROS\,2~\cite{macenski2022ros2} is not purely mechanical: ROS\,2 replaces ROS\,1's implicit, lossless transport with an explicit QoS contract, and modern distributions bundle substantially newer \texttt{oneTBB} and Eigen than the code was written against.
Either change can break a tightly-coupled filter silently -- the port compiles, runs, and then diverges.
All estimator math (voxel map, GICP + point-to-plane fusion, iterated-EKF propagation) is unchanged from upstream~\cite{chen2024iglio,segal2009gicp}; the migration itself (\texttt{catkin} to \texttt{ament\_cmake}, \texttt{roscpp} to \texttt{rclcpp}, \texttt{.launch} to \texttt{.launch.py}, \texttt{tf} to \texttt{tf2\_ros}) is mechanical and not detailed here.
This report centers on debugging the two failures that surface only after that migration (Section~\ref{sec:numerical}), plus corrected sensor parsing (Section~\ref{sec:sensors}) and YAML-configurable QoS/TF/output paths (Section~\ref{sec:usability}).
The estimator and its performance claims belong to the original authors~\cite{chen2024iglio}.

% ===========================================================================
\section{Diagnosing Numerical Failures}
\label{sec:numerical}
% ===========================================================================
After a mechanically correct port -- math unchanged -- the solver died with \texttt{NaN} Hessian matrix entries and aborts inside \texttt{SO3::exp(nan)}.
Two distinct, environment-level root causes were isolated; they are the most broadly transferable outcome of this work.

\subsection{A QoS mismatch silently dropping IMU samples}
\label{sec:qos}
The initial port subscribed using \texttt{rclcpp::SensorDataQoS()}, i.e.\ \texttt{BEST\_EFFORT} with \texttt{KEEP\_LAST(5)}, while the data source (bag or driver) published \texttt{RELIABLE} with a deeper queue.
This pairing is compatible but silently degrades to best-effort: under the single-threaded \texttt{spin\_some} + processing loop, backed-up IMU samples were dropped and reordered, corrupting the signal the iterated filter's propagation depends on.
The fix subscribes \texttt{RELIABLE} with a deep IMU queue (\texttt{KeepLast(2000)}) and \texttt{KeepLast(10)} for the LiDAR -- reliability is now a YAML parameter (Section~\ref{sec:usability}) -- plus a guard rejecting non-physical prediction time steps ($\Delta t \le 0$ or $\Delta t > 0.5$\,s).

\paragraph{Lesson.} ROS\,1 had no QoS, so transport correctness was implicit.
In ROS\,2, QoS is a concern for tightly-coupled estimators, not just throughput.

\subsection{An uninitialized \texttt{oneTBB} + Eigen reduction accumulator}
\label{sec:tbb}
The true crash lay in constraint assembly.
\texttt{ConstructGICPConstraints} and \texttt{ConstructPoint2PlaneConstraints} assembled the normal equations with \texttt{tbb::parallel\_reduce}, using a fixed-size \texttt{Eigen::Matrix} as the reduction Value type.
The \texttt{oneTBB} shipped with Jazzy default-constructs the split accumulators it joins in parallel -- and a fixed-size Eigen matrix's default constructor does not zero-initialize its storage.
Uninitialized memory was summed into $H$ and $b$, yielding \texttt{NaN} Hessian matrixes; GCC pinpointed an SSE register used uninitialized inside the TBB join body.

An interim fix split each constructor into a parallel \texttt{tbb::parallel\_for} correspondence search feeding a serial $H$/$b$ assembly loop -- correct, but a few milliseconds slower per scan, visible as cadence gaps at 4$\times$ bag playback.
The final fix instead keeps upstream's single fused \texttt{tbb::parallel\_reduce} and changes only the reduction Value type: the bare \texttt{Eigen::Matrix} becomes a small accumulator struct whose members are zero-initialized in-class (\texttt{Eigen::Matrix::Zero()}) and joined via \texttt{operator+}, so every split TBB creates -- including the identity value -- starts zeroed.
This is the minimal correct change: structurally identical to upstream, fully parallel, and free of the uninitialized-read fault.

\paragraph{Lesson.} The fault is not that \texttt{tbb::parallel\_reduce} is unsafe -- a bare fixed-size Eigen matrix used as the reduce Value is left uninitialized by oneTBB's default-constructed splits.
Wrapping it in a zero-initializing accumulator struct preserves the parallel reduction rather than forfeiting it.
The bug is invisible on the toolchain the code was written against and surfaces only on the newer TBB bundled with modern ROS\,2.

% ===========================================================================
\section{Sensor Support}
\label{sec:sensors}
% ===========================================================================
The Ouster point struct/parser was updated to the current \texttt{PointCloud2} field layout in revision 7 sensors, restoring motion de-skewing that was broken for these newer sensors in the original code.
A dedicated point type (\texttt{lidar\_type: M1600}) adds Velodyne Velarray~M1600 support, carried forward from an intermediate fork by Pedro Tomás\footnote{\url{https://github.com/pedrotomas27/ig_lio}}.
Only the \texttt{ouster} and \texttt{livox\_points} data paths are validated on hardware (\texttt{velodyne}, \texttt{M1600}, \texttt{Hesai} and \texttt{livox} are ported but unverified).

\paragraph{Livox: two independent ingestion paths.} Livox sensors publish either a driver-specific \texttt{CustomMsg} or, in an alternative transfer mode (on newer sensors such as the Mid-360), standard \texttt{sensor\_msgs/PointCloud2}.
The \texttt{CustomMsg} path has been made optional and compile-time gated: \texttt{CMakeLists.txt} quietly tries to find the Livox driver, deliberately absent from the package manifest so dependency resolution succeeds without it; when found, a \texttt{HAVE\_LIVOX} macro compiles in the guarded parser and subscription, ported from upstream; otherwise the build proceeds with Livox disabled and selecting \texttt{lidar\_type: livox} without the driver fails fast with an actionable message.
The \texttt{PointCloud2} path (\texttt{lidar\_type: livox\_points}) needs none of this gating, since it is handled by the same \texttt{PointCloud2} subscription as the other sensors, via a dedicated point type/parser, and is therefore always available.
Together, these mean users on \texttt{PointCloud2}-publishing Livox sensors never need the Livox driver.

% ===========================================================================
\section{Usability and Validation}
\label{sec:usability}
\begin{figure}[ht]
\centering
\includegraphics[width=0.68\linewidth]{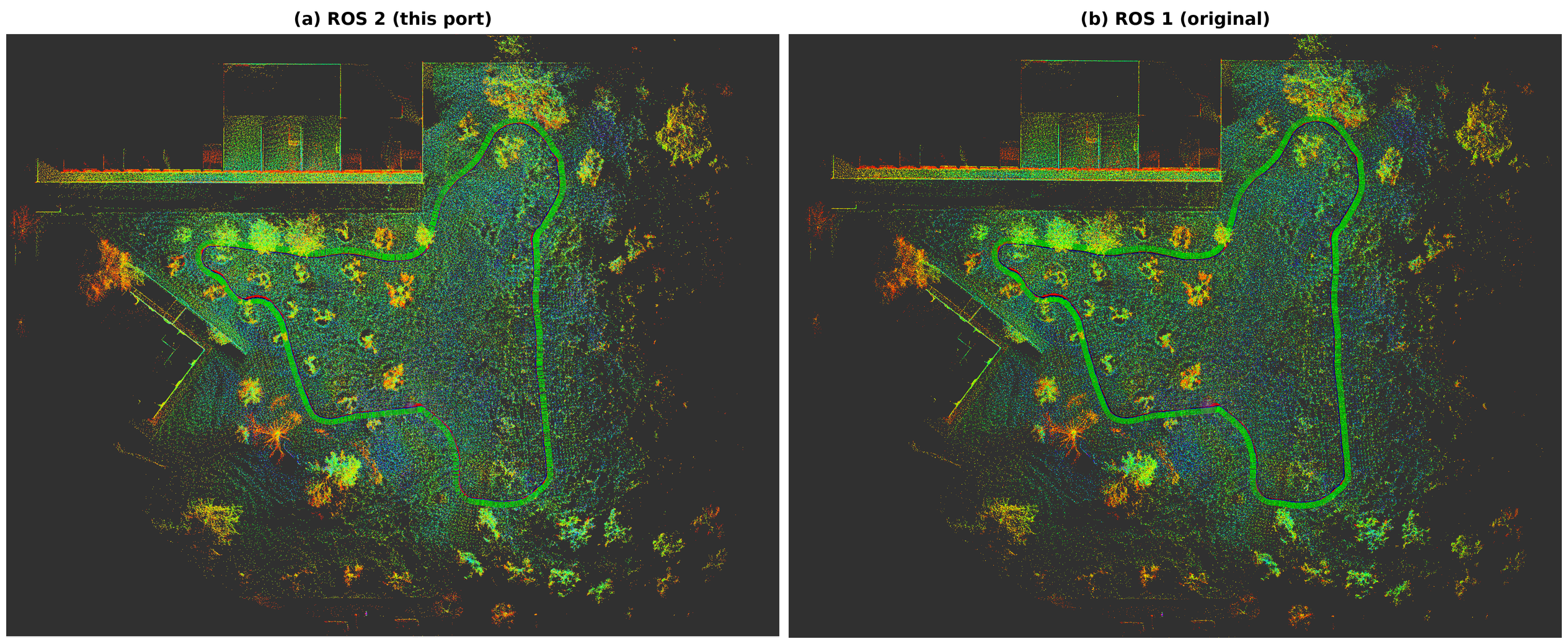}
\caption{Estimated trajectory in brighter green over the accumulated point-cloud
map, for the same sequence run on (a) this ROS\,2 port and (b) the original
ROS\,1 \mbox{iG-LIO}.}
\label{fig:ros1-vs-ros2}
\end{figure}

% ===========================================================================
Beyond the fixes above, the port exposes \texttt{qos\_reliability} (\texttt{best\_effort}$|$\texttt{reliable}), \texttt{odom\_frame}/\texttt{base\_frame} names, and \texttt{result\_directory} via YAML; publishes the estimator's actual odometry covariance under a REP-105-compliant \texttt{odom} frame; and writes the trajectory in TUM format to a path that persists across rebuilds, directly loadable by \texttt{evo}~\cite{evo} for APE/RPE evaluation.
As a qualitative sanity check, Figure~\ref{fig:ros1-vs-ros2} overlays the estimated trajectory on the accumulated point-cloud map for the same sequence run on this ROS\,2 port and on the original ROS\,1 codebase: the two are qualitatively identical, indicating that the migration and numerical-robustness fixes of Section~\ref{sec:numerical} preserve the estimator's behavior.

% ===========================================================================
\section{Conclusion}
\label{sec:conclusion}
% ===========================================================================
A faithful migration of a iG-LIO was not numerically stable by default.
Two failures -- an IMU-dropping QoS mismatch and an uninitialized \texttt{oneTBB}+Eigen reduction accumulator -- arose from the modern toolchain and broke the filter silently despite unchanged mathematics; both fixes are small but generalize: treat QoS as a correctness contract, and never reduce over an uninitialized fixed-size Eigen accumulator.
If you use this ROS\,2 port, please cite both the original \mbox{iG-LIO} paper~\cite{chen2024iglio} and this report.

% ===========================================================================
\section*{Acknowledgments}
% ===========================================================================
\begin{sloppypar}
Supported by FCT under the Affiliated Ph.D.\ CMU Portugal Program (ref.\ PRT/BD/153920/2022), the ISR's Multiannual Funding for R\&D Units (Project UIDB/00048/2020), and the ForestSphere project (Grant 2025.01496.DT4ST, DT4ST programme, PRR/FCT, with ARTE).
We thank the original \mbox{iG-LIO} authors and Pedro Tomás, who authored the intermediate fork integrating the Velarray~M1600 work.
\end{sloppypar}

\bibliographystyle{ieeetr}
\bibliography{references}

@article{chen2024iglio,
  author  = {Chen, Zijie and Xu, Yong and Yuan, Shenghai and Xie, Lihua},
  title   = {{iG-LIO}: An Incremental {GICP}-Based Tightly-Coupled {LiDAR}-Inertial Odometry},
  journal = {IEEE Robotics and Automation Letters},
  year    = {2024},
  volume  = {9},
  number  = {2},
  pages   = {1883--1890},
  doi     = {10.1109/LRA.2024.3349915}
}

@article{macenski2022ros2,
  author  = {Macenski, Steven and Foote, Tully and Gerkey, Brian and Lalancette, Chris and Woodall, William},
  title   = {Robot Operating System 2: Design, Architecture, and Uses in the Wild},
  journal = {Science Robotics},
  year    = {2022},
  volume  = {7},
  number  = {66},
  pages   = {eabm6074},
  doi     = {10.1126/scirobotics.abm6074}
}

@inproceedings{segal2009gicp,
  author    = {Segal, Aleksandr and Haehnel, Dirk and Thrun, Sebastian},
  title     = {Generalized-{ICP}},
  booktitle = {Robotics: Science and Systems (RSS)},
  series    = {RSS},
  year      = {2009},
  number    = {4},
  pages     = {435}
}

@misc{evo,
  author       = {Grupp, Michael},
  title        = {{evo}: Python Package for the Evaluation of Odometry and {SLAM}},
  year         = {2017},
  howpublished = {\url{https://github.com/MichaelGrupp/evo}}
}

@misc{onetbb,
  author       = {{Intel Corporation}},
  title        = {{oneAPI} Threading Building Blocks ({oneTBB})},
  howpublished = {\url{https://github.com/uxlfoundation/oneTBB}}
}

\end{document}